\documentclass{article}

\usepackage{soul}
\usepackage[linesnumbered]{algorithm2e}
\usepackage{todonotes}
\usepackage{url}
\usepackage{amsmath,amssymb,amsthm,amsfonts}
\usepackage{booktabs}
\usepackage{amsfonts}
\usepackage{caption}
\usepackage{subcaption}
\usepackage{mathbbol}
\usepackage{multirow}
\usepackage{graphicx, array, blindtext}
\graphicspath{{./Figures/}}
\usepackage{amsmath}
\usepackage{sidecap}
\usepackage{pbox}
\usepackage{arydshln}
\usepackage[numbers]{natbib}
\usepackage{floatrow}
\newfloatcommand{capbtabbox}{table}[][\FBwidth]
\usepackage{blindtext}
\newcommand{\bs}{\boldsymbol{s}}

\newcommand{\bu}{\boldsymbol{u}}
\newcommand{\bv}{\boldsymbol{v}}

\newtheorem{conjecture}{Conjecture}

\newtheorem{proposition}{Proposition}

\def\defterm#1{\textbf{#1}}
\def\bs#1{\boldsymbol{#1}}
\def\set#1{\bs{#1}}

\def\reals{\mathbb{R}}

\newcommand{\graph}{\G}

\newcommand{\nodes}{V}

\newcommand{\numnodes}{N}
\newcommand{\unode}{u}
\newcommand{\vnode}{v}

\newcommand{\edges}{E}
\newcommand{\relation}{R}

    \newcommand{\motif}{\mu}
  
    \newcommand{\mmotif}{\set{M}}

\newcommand{\gprob}{p}
\newcommand{\prior}{p}
\newcommand{\linstance}{\set{z}}
\newcommand{\ldim}{t}

\newcommand{\nodem}{i}
\newcommand{\noden}{j}
\newcommand{\G}{G}
\newcommand{\A}{\set{A}} 
 
\newcommand{\Z}{\set{Z}}
 
\newcommand{\prob}[1]{\Tilde{#1}}

\newcommand{\Featuredim}{l}

\newcommand{\featureFunction}{\phi}

\usepackage[utf8]{inputenc} 
\usepackage[T1]{fontenc}    
\usepackage{url}            
\usepackage{booktabs}       
\usepackage{amsfonts}       
\usepackage{nicefrac}       
\usepackage{microtype}      

\title{Computing Expected Motif Counts for Exchangeable Graph Generative Models}
\author{%
   Oliver Schulte\\
 School of Computing Science, Simon Fraser University\\
 Vancouver, Canada
 }

\begin{document}
\date{\today}
\maketitle

\begin{abstract}
Estimating the expected value of a graph statistic is an important inference task for using and learning graph models. This note presents a scalable estimation procedure for expected motif counts, a widely used type of graph statistic. The procedure applies for generative mixture models of the type used in neural and Bayesian approaches to graph data. 
\end{abstract}



\section{Introduction and Problem Definition}

A graph is a pair $\G=(V,E)$ comprising a finite set of $\numnodes$ nodes and edges.
The edges can be represented by an indicator function $\relation_{\G}: \nodes^{2} \rightarrow \{0,1\}$ such that $\relation_{G}(\unode,\vnode)=1$ if $(\unode,\vnode)\in E$, and 0 otherwise. Given a node ordering, a graph can be represented by an adjacency matrix $\A_{\numnodes \times \numnodes}$.

A \defterm{descriptor function} $\featureFunction$ maps a graph $\G$ to a $\Featuredim$-dimensional \defterm{graph statistic} such that $\featureFunction(\G)  \in \reals^{\Featuredim}$~\cite{o2021evaluation}. In the following we consider a probability distribution $\gprob$ over graphs of a fixed size $\numnodes$. The \defterm{expected graph statistic} vector is given by

\begin{equation} \label{eq:xstat}
    E[\featureFunction] = \sum_{\G} \gprob(\G) \featureFunction(\G).
\end{equation}

{\em The problem is to compute the expected graph statistic for a given distribution $\gprob$ and graph descriptor $\featureFunction$.} This note addresses the case where $\gprob$ is a mixture of graph distributions with conditionally independent links, and $\featureFunction$ is a graph motif. Briefly, we show that under these assumptions, the expected graph statistic can be estimated efficiently in two steps. (1) As is known from previous work, variational inference can be used to approximate the posterior of the mixture variable $\Z$ with few samples~\cite{kipf2016variational}. (2) Our main result shows that given a mixture sample $\linstance$, the expected graph statistic can be computed 
by applying the graph descriptor to a single matrix, the expected adjacency matrix conditional on $\linstance$. Since the links are conditionally independent given $\linstance$, finding the expected adjacency matrix takes linear time in the size of the matrix. The main steps in the argument for (2) are as follows.

\begin{enumerate}
    \item A motif can be represented as a sum of products of binary link assignments. 
    \item Given (conditionally) independent links, the expected value of a product of link assignments is the product of expected values. The expected adjacency matrix entries contain the expected values for each link assignment.
    \item Since the expectation of a sum is the sum of expectations, computing the motif instance sum in the expected adjacency matrix gives the expectation of the sum.

\end{enumerate}

Computing the expected motif count has several applications in machine learning, for example:  (1) Assessing the statistical significance of a motif in an observed network by comparing the expected and observed counts~\cite{martorana2020establish}. (2) Training a generative graph model with a moment-matching objective to minimize the difference between observed and expected counts~\cite{zahirnia2022micro}. The work of~\citet{zahirnia2022micro} shows that for a deep graph generative model, the expected adjacency matrix can be found efficiently, and presents several procedures for computing common statistics from the expected adjacency matrix. Their work, however, does not show that the statistics computed from the expected adjacency matrix represent the expected model statistics, which is implied by our result for motif counts. 

\section{Mixture  Graph Distributions}
Let $\Z \in \reals^{\ldim}$ be a latent variable with prior distribution $p(\Z)$. A decoder deterministically maps a sample $\linstance$ to a weighted graph $\prob{\G_{\linstance}} = (\nodes,\prob{\relation}_{\linstance})$ where $\prob{\relation}_{\linstance}: \nodes^{2} \rightarrow [0,1]$ gives the probability that a link exists between any pair of nodes, and different link probabilities are independent of each other. 
The resulting mixture model is the following. 


\begin{align}
\gprob(\G) = \int P(\G|\prob{\G}_{\linstance}) \prior(\linstance) d\linstance 
\label{eq:mixture}\\ 
  P(\G|\prob{\G}_{\linstance}) =
     \prod_{\unode \in \nodes} \prod_{\vnode \in \nodes} \prob{\relation}_{\linstance}(\unode,\vnode)^{\relation_{\G}(\unode,\vnode)} (1-\prob{\relation}_{\linstance}(\unode,\vnode))^{1-\relation_{\G}(\unode,\vnode)}. \notag
\end{align}

 A generalization of deFinetti's exchangeability theorem to infinite matrix data states that all permutation-invariant (exchangeable) distributions $\gprob$ over infinite graphs can be represented as a mixture of the form~\eqref{eq:mixture}~\cite{orbanz2014bayesian}. A similar representation theorem can be established for exchangeable probability distributions over finite graphs under the projectivity assumption~\cite{Jaeger2020}. Intuitively, projectivity means that the probability of a subgraph does not depend on the population size (i.e., the marginal probability of a subgraph $\graph^{m}$ comprising $m$ nodes is the same for any node set size $n \geq m$). 
 
 \section{Motifs}
 
\begin{figure}
\begin{floatrow}
\ffigbox{%
  \includegraphics{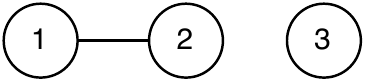}%
}{%
  \caption{A motif template graph\label{fig:motif}}%
}
\capbtabbox{%
  \begin{tabular}{|l|l|l|l|}
\hline
  & 1 & 2 & 3 \\ \hline
1 & 0 & 1 & 0 \\ \hline
2 & 1 & 0 & 0 \\ \hline
3 & 0 & 0 & 0 \\ \hline
\end{tabular}
}{%
  \caption{The motif adjacency matrix
\label{table:motif-matrix}}%
}
\end{floatrow}
\begin{floatrow}
\ffigbox{%
  \includegraphics{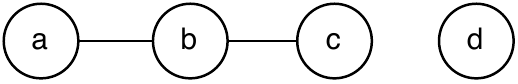}%
}{%
  \caption{An input graph \label{fig:input-graph}}%
}
\capbtabbox{%
 \begin{tabular}{|c|c|c|}
\hline
\multicolumn{1}{|l|}{$v_1$} & \multicolumn{1}{l|}{$v_2$} & \multicolumn{1}{l|}{$v_3$} \\ \hline \hline
a                           & b                          & d                          \\ \hline
b                           & a                          & d                          \\ \hline
b                           & c                          & d                          \\ \hline
c                           & b                          & d                          \\ \hline
\end{tabular}%
}{%
  \caption{The Motif Count in the example motif and input graph. 
\label{table:motif-count}}%
}
\end{floatrow}
\end{figure}

 Intuitively, a motif specifies a small subgraph; a motif count for a graph specifies how many times the motif graph appears in the larger graph. A motif can be visualized as an ordered template graph (see Figure~\ref{fig:motif}). Formally, a motif of arity $k$ can be represented by an $k\times k$ adjacency matrix $\mmotif$ with generic entry $\mmotif[\nodem,\noden]$ (see Table~\ref{table:motif-matrix}).

 The \defterm{motif
 indicator function} takes as input a graph and an ordered list of $k$ nodes from a fixed node set $\nodes$,  and returns 1 if the ordered subgraph induced by the $k$ nodes matches the motif. The motif indicator function can be computed by the following \defterm{product formula}.
 
 \begin{equation} \label{eq:mcount}
     \motif(\G,\langle\vnode_1,\ldots,\vnode_k \rangle) = \prod_{\nodem=1}^{k} \prod_{\noden=1}^{k}  \relation_{\G}(\vnode_\nodem,\vnode_\noden)^{\mmotif[\nodem,\noden]}  \cdot (1-\relation_{\G}(\vnode_\nodem,\vnode_{\noden}))^{(1-\mmotif[\nodem,\noden]}) 
 \end{equation}
 where each  $\vnode_i$ is in the domain $\nodes$ (see Table~\ref{table:motif-matrix}). The \defterm{motif count} in a graph is given by
 
\begin{equation}
   \featureFunction_{\motif}(\G) = \sum_{\set{\vnode} \in \nodes^{k}}  \motif(\G,\set{\vnode}).
\end{equation}

Table~\ref{table:motif-count} illustrates the motif count. An undirected edge is equivalent to two pairs of directed edges.

Note that Equation~\eqref{eq:mcount} naturally extends to a weighted graph $\prob{\G}= (\nodes,\prob{R})$: the expression $\prob{\relation}(\vnode_\nodem,\vnode_\noden)^{\mmotif[\nodem,\noden]}  \cdot (1-\prob{\relation}(\vnode_\nodem,\vnode_{\noden}))^{(1-\mmotif[\nodem,\noden]})$ can be read as ``if the template graph specifies a node between links $\nodem$ and $\noden$, return the weight  $\prob{\relation}(\vnode_\nodem,\vnode_\noden)$; otherwise return the weight $(1-\prob{\relation}(\vnode_\nodem,\vnode_{\noden}))$''. 
We write $\featureFunction_{\motif}(\prob{\G}) = \sum_{\set{\vnode} \in \nodes^{k}}  \motif(\prob{\G},\set{\vnode})$ for the motif count in a weighted graph. 
We next consider how to compute the expected motif count.

\section{Expected Motif Counts for Mixture Models}

The expected motif count for a mixture model can be computed as the mixture of expected motif counts:

\begin{align}
E[\featureFunction_{\motif}]  = \sum_{\G} \gprob(\G) \featureFunction_{\motif}(\G) = \sum_{\G} \int P(\G|\prob{\G}_{\linstance}) \prior(\linstance) d\linstance \featureFunction_{\motif}(\G) \notag\\
= \int  \sum_{\G} P(\G|\prob{\G}_{\linstance}) \featureFunction_{\motif}(\G) \prior(\linstance) d\linstance = 
 \int    E[\featureFunction_{\motif}|\linstance] \prior(\linstance) d\linstance \label{eq:interchange}
\end{align}

where Equation~\eqref{eq:interchange} follows from changing the order of integration and summations. The inner sum of Equation~\eqref{eq:interchange} is the expected value of the statistic conditional on an embedding $\linstance$, and the integral the expectation of the sum over the latent space. Given an efficient way to evaluate the sum, the integral can be approximated by sampling $\linstance$-values from the prior $p(\linstance)$. Variational inference can be used to reduce the number of samples required~\cite{kipf2016variational}. The next proposition provides a closed form expression for computing the expectation. 

\begin{proposition} \label{prop:main}
For each motif $\motif$ and latent value $\linstance$, the expected motif count equals the motif count computed from the expected graph:
\begin{equation}
E[\featureFunction_{\motif}|\linstance] = \featureFunction_{\motif}(\prob{\G}_{\linstance})
\end{equation}

\end{proposition}

Since links are independent given $\linstance$, the graph $\prob{\G}_{\linstance}$ is the expectation over link indicator variables $\prob{\relation}_{\linstance}$. Given a node ordering, the expectation over the binary matrices $\A$ representing unweighted graphs can be computed from the expected adjacency matrix $\prob{\set{A}}$, which represents the weighted graph $\prob{\G}_{\linstance}$.

\paragraph{Matrix View}  In terms of adjacency matrices, the essence of the proof of Proposition~\ref{prop:main} is that, when links are independent, the expectation of an adjacency matrix product is the product of the expected adjacency matrices. This means that if the motif count is defined in terms of matrix summation and multiplication, the expected motif count can be computed by applying the motif count operation to the expected adjacency matrix. 

For example, the number of triangles in an undirected graph can be counted as the number of length-three paths that start and end at a node $\nodem$: 

\begin{equation*}
    \motif_{T}(\A) = \sum_{i}\A^3[i,i]
\end{equation*}

Interchanging expectations with sums and products, we have that

\begin{equation*}
    E[\motif_{T}|\linstance] = E[\sum_{i}\A^3[i,i]|\linstance] = \sum_{i}E[\A^3[i,i]|\linstance] = \sum_{i}(\prob{\A}_{\linstance}^3)[i,i] = \motif_{T}(\prob{\A}_{\linstance}). 
\end{equation*}

\section{Ordered vs. Unordered Motifs} Proposition~\ref{prop:main} is valid for ordered motifs, which are satisfied by a {\em tuple} of nodes. Defining subgraphs in terms of  tuples that satisfy them is natural from the point of view of relational query languages like SQL and the domain relational calculus, where the answer to a query is a set of tuples that satisfy the query~\cite{Ramakrishnan2003}. The domain relational calculus shows how first-order logic can be used as an expressive for defining queries and also motifs. For example, the motif of Figure~\ref{fig:motif} can be defined by the formula

\begin{equation*} \label{eq:fol}
    R(X_1,X_2), \neg R(X_2,X_3), \neg R(X_1,X_3)
\end{equation*}

where $X_1,X_2,X_3$ are first-order variables (not random variables) that are instantiated by individual nodes as in a template or a plate model. Intuitively, Formula~\ref{eq:fol} can be read as ``for any nodes $x_1,x_2,x_3$, they satisfy the motif if $x_1$ links to $x_2$ and neither $x_2$ nor $x_3$ links to $x_1$.''

It is also possible to define motif counts for {\em unordered} sets of nodes, where a set of nodes $\{\vnode_1,\ldots,\vnode_{k}\}$ satisfies a motif in a graph $\G$ if the induced subgraph is isomorphic to the motif graph~\cite{bouritsas2022improving}. We show that expected instantiation counts for the set-based definition are related to expected instantiation counts for the tuple-based definition by a constant that depends on the motif but not on the mixture distribution.


Let $\motif$ be a motif of arity $k$, let $\G= (\nodes,\edges)$ be a graph, and suppose that $U \subseteq \nodes$ is a subset of nodes of size $k$. Define the set instantiation count as follows.

\begin{align*}
    \overline{\featureFunction}_{\motif}(\G)= \sum_{U \subseteq \nodes, |U|=k}  \overline{\motif}(\G,U)\\
    \overline{\motif}(\graph,U) =
    \begin{cases}1, & \text{if there is an ordering $\set{u} = \langle u_1,\ldots,u_k \rangle$  of $U$ s.t. $\motif(\graph,\set{u})=1$} \\
    0, & \text{otherwise}
    \end{cases}
\end{align*}

In the example of Table~\ref{table:motif-matrix}, there are two sets that satisfy the motif, namely $\{a,b,d\}$ and $\{b,c,d\}$. Therefore $\overline{\featureFunction}_{\motif}(\G) = 2$. In the example, each set instance gives rise to two tuple instances. 
The next proposition states that for any input graph $\G$, the number of tuple instantiations of a motif is the number of set instantiations, multiplied by the number of automorphisms of the motif graph. 

A graph \defterm{automorphism} is a 1-1 mapping of the vertices onto itself that preserves edges. For an adjacency matrix $\mmotif_{k \times k}$, such as a motif adjacency matrix (see Table~\ref{table:motif-matrix}), an automorphism is a permutation $\pi$ of the index set $\{1,\ldots,k\}$ such that for all $\nodem,\noden$ we have $\mmotif[\nodem,\noden] = \mmotif[\pi(\nodem),\pi(\noden)] $.  

In the example of Figure~\ref{fig:motif}, the permutation $\pi(1) = 3, \pi(2) = 1, \pi(3)=3$ is an automorphism. Together with the identity permutation, the motif graph in this example therefore admits two automorphisms. 

\begin{conjecture} \label{prop:automorphism}
    Let $\motif$ be a motif admitting $\mathrm{Aut}(\mu)$ automorphisms. 
    \begin{enumerate}
        \item For all graphs $\graph$ we have $\featureFunction_{\motif}(\G)= \mathrm{Aut}(\mu) \times \overline{\featureFunction}_{\motif}(\G)$.
        \item $E[\featureFunction_{\motif}] =\mathrm{Aut}(\mu) \times E[\overline{\featureFunction}_{\motif}] $
    \end{enumerate}
\end{conjecture} 

We believe that this result is well-known in the community (see~\cite[Appendix C]{bib:arxiv-version}),
but have not been able to find an explicit proof in the literature. The 
conjecture implies that the efficient method for computing tuple motif counts provided by Proposition~\ref{prop:main} can be extended to set motif counts, given the number of automorphisms of the motif graphs. For small graphs like motif graphs, the number of automorphisms can be found quickly by enumeration~\cite{grochow2007network}. 








\section{Conclusion}

Computing expected motif counts is a useful computational task for network modelling. This note provided an efficient new approach for an important model class---mixtures of models with independent links---which is widely used in deep graph learning and Bayesian analysis of graph data. We showed that conditional on latent features (embedings) that render links conditionally independent, the expected motif count is the motif count of the expected graph. It can therefore be computed exactly given latent features, without the need for generating simulated networks, at the computational cost of finding the expected graph. The only sampling required is sampling latent features. 

\section*{Acknowledgements} This research was supported by a discovery grant form the Natural Sciences and Engineering Research Council of Canada. Manfred Jaeger and Abdolreza Mirzaei provided valuable comments on a draft of this paper. 

\section*{Proof of Proposition~\ref{prop:main}.}

\begin{proof}
For a fixed tuple of nodes $\set{\vnode}$, define the following random variables.

\begin{itemize}
\item $r^{\set{\vnode}}_{\nodem\noden}$ returns $\relation_{\G}(\vnode_\nodem,\vnode_\noden)$ (i.e., if 1 if the link exists, 0 otherwise).
\item $\delta^{\set{\vnode}}_{\nodem\noden} = (r^{\set{\vnode}}_{\nodem\noden})^{\mmotif[\nodem,\noden]}  \cdot (1-r^{\set{\vnode}}_{\nodem\noden})^{(1-\mmotif[\nodem,\noden)})$ 
\end{itemize}

Since the $r^{\set{\vnode}}_{\nodem\noden}$ are independent given $\linstance$, so are the $\delta^{\set{\vnode}}_{\nodem\noden}$ variables. If $\mmotif[\nodem,\noden] = 1$, then $E[\delta^{\set{\vnode}}_{\nodem\noden}] = \prob{\relation}_{\linstance}(\vnode_\nodem,\vnode_\noden)$. If $\mmotif[\nodem,\noden] = 0$, then $E[\delta^{\set{\vnode}}_{\nodem\noden}] = (1-\prob{\relation}_{\linstance}(\vnode_\nodem,\vnode_\noden))$. Therefore

\begin{equation*}
    E[\delta^{\set{\vnode}}_{\nodem\noden}] = \prob{\relation}_{\linstance}(\vnode_\nodem,\vnode_\noden)^{\mmotif[\nodem,\noden]}  \cdot (1-\prob{\relation}_{\linstance}(\vnode_\nodem,\vnode_\noden)^{(1-\mmotif[\nodem,\noden])}).
\end{equation*}

Considering the expected motif count, we now have the following. 

\begin{align}
    E[\featureFunction_{\motif}|\linstance] = E[\sum_{\set{\vnode} \in \nodes^{k}} \prod_{\nodem=1}^{k} \prod_{\noden=1}^{k} \delta^{\set{\vnode}}_{\nodem \noden}] = \sum_{\set{\vnode} \in \nodes^{k}} E[ \prod_{\nodem=1}^{k} \prod_{\noden=1}^{k} \delta^{\set{\vnode}}_{\nodem \noden}] \notag\\
    = \sum_{\set{\vnode} \in \nodes^{k}} \prod_{\nodem=1}^{k} \prod_{\noden=1}^{k} E[\delta^{\set{\vnode}}_{\nodem \noden}] \label{eq:ind-product} \\
    = \sum_{\set{\vnode} \in \nodes^{k}} \prod_{\nodem=1}^{k} \prod_{\noden=1}^{k} \prob{\relation}_{\linstance}(\vnode_\nodem,\vnode_\noden)^{\mmotif[\nodem,\noden]}  \cdot (1-\prob{\relation}_{\linstance}(\vnode_\nodem,\vnode_\noden)^{(1-\mmotif[\nodem,\noden])}) \notag\\
    = \featureFunction_{\motif}(\prob{\G}_{\linstance}) \notag
\end{align}

Line~\eqref{eq:ind-product} follows because the expectation of a product of independent random variables is the product of their expectations. 
\end{proof}

\bibliography{main.bib,master.bib}

\begin{thebibliography}{10}
\providecommand{\natexlab}[1]{#1}
\providecommand{\url}[1]{\texttt{#1}}
\expandafter\ifx\csname urlstyle\endcsname\relax
  \providecommand{\doi}[1]{doi: #1}\else
  \providecommand{\doi}{doi: \begingroup \urlstyle{rm}\Url}\fi

\bibitem[Bouritsas et~al.(2022)Bouritsas, Frasca, Zafeiriou, and
  Bronstein]{bouritsas2022improving}
Giorgos Bouritsas, Fabrizio Frasca, Stefanos~P Zafeiriou, and Michael
  Bronstein.
\newblock Improving graph neural network expressivity via subgraph isomorphism
  counting.
\newblock \emph{IEEE Transactions on Pattern Analysis and Machine
  Intelligence}, 2022.

\bibitem[Grochow and Kellis(2007)]{grochow2007network}
Joshua~A Grochow and Manolis Kellis.
\newblock Network motif discovery using subgraph enumeration and
  symmetry-breaking.
\newblock In \emph{Annual International Conference on Research in Computational
  Molecular Biology}, pages 92--106. Springer, 2007.

\bibitem[Jaeger and Schulte(2020{\natexlab{a}})]{Jaeger2020}
Manfred Jaeger and Oliver Schulte.
\newblock A complete characterization of projectivity for statistical
  relational models.
\newblock In Christian Bessiere, editor, \emph{Proceedings {IJCAI-20}}, pages
  4283--4290. International Joint Conferences on Artificial Intelligence
  Organization, 7 2020{\natexlab{a}}.
\newblock \doi{10.24963/ijcai.2020/591}.
\newblock URL \url{https://doi.org/10.24963/ijcai.2020/591}.
\newblock Main track.

\bibitem[Jaeger and Schulte(2020{\natexlab{b}})]{bib:arxiv-version}
Manfred Jaeger and Oliver Schulte.
\newblock A complete characterization of projectivity for statistical
  relational models.
\newblock \emph{arXiv preprint arXiv:2004.10984}, 2020{\natexlab{b}}.

\bibitem[Kipf and Welling(2016)]{kipf2016variational}
Thomas Kipf and M.~Welling.
\newblock Variational graph auto-encoders.
\newblock \emph{ArXiv}, abs/1611.07308, 2016.

\bibitem[Martorana et~al.(2020)Martorana, Micale, Ferro, and
  Pulvirenti]{martorana2020establish}
Emanuele Martorana, Giovanni Micale, Alfredo Ferro, and Alfredo Pulvirenti.
\newblock Establish the expected number of induced motifs on unlabeled graphs
  through analytical models.
\newblock \emph{Applied Network Science}, 5\penalty0 (1):\penalty0 1--23, 2020.

\bibitem[O'Bray et~al.(2022)O'Bray, Horn, Rieck, and
  Borgwardt]{o2021evaluation}
Leslie O'Bray, Max Horn, Bastian Rieck, and Karsten Borgwardt.
\newblock Evaluation metrics for graph generative models: Problems, pitfalls,
  and practical solutions.
\newblock In \emph{International Conference on Learning Representations}, 2022.

\bibitem[Orbanz and Roy(2014)]{orbanz2014bayesian}
Peter Orbanz and Daniel~M Roy.
\newblock Bayesian models of graphs, arrays and other exchangeable random
  structures.
\newblock \emph{IEEE transactions on pattern analysis and machine
  intelligence}, 37\penalty0 (2):\penalty0 437--461, 2014.

\bibitem[Ramakrishnan and Gehrke(2003)]{Ramakrishnan2003}
Raghu Ramakrishnan and Johannes Gehrke.
\newblock \emph{Database Management Systems}.
\newblock McGraw-Hill, 3rd edition, 2003.

\bibitem[Zahirnia et~al.(2022)Zahirnia, Schulte, Naddaf, and
  Li]{zahirnia2022micro}
Kiarash Zahirnia, Oliver Schulte, Parmis Naddaf, and Ke~Li.
\newblock Micro and macro level graph modeling for graph variational
  auto-encoders.
\newblock \emph{arXiv preprint arXiv:2210.16844}, 2022.

\end{thebibliography}

\end{document}